\documentclass[letterpaper]{article} % DO NOT CHANGE THIS
\usepackage[preprint]{aaai2027}
\usepackage[hyphens]{url} % DO NOT CHANGE THIS
\usepackage{graphicx} % DO NOT CHANGE THIS
\usepackage{natbib} % DO NOT CHANGE THIS AND DO NOT ADD OPTIONS
\usepackage{caption} % DO NOT CHANGE THIS AND DO NOT ADD OPTIONS
\usepackage{booktabs}
\usepackage{amsmath,amssymb}
\usepackage{multirow}
\usepackage{subcaption}

\title{MASQ: Mask-Aware Spatiotemporal Quantization for Unsupervised Skeleton Action Segmentation}
\author{
    Xinyao Qin\equalcontrib,
    Linxiang Peng\equalcontrib,
    Youbao Ye,
    Di Yang\corresponding,
    Jiangtao Wang\corresponding
}

\affiliations{
University of Science and Technology of China, China\\
\{qxy15196216301,plx,yybao\}@mail.ustc.edu.cn,
\{di.yang,wangjiangtao@\}@ustc.edu.cn
}

\begin{document}

\maketitle

\begin{abstract}
Unsupervised skeleton-based temporal action segmentation is a crucial task for understanding human behavior in long untrimmed sequences. Recent approaches often rely on discrete quantization to discover action boundaries from motion representations. However, when spatial masking is introduced for representation learning, it can introduce representation ambiguity, while discrete quantization further amplifies small fluctuations in the latent space. The interaction between these two factors often leads to unstable code switching and severe temporal jitter near action boundaries.
To address these limitations, we propose a novel Mask-aware Action Spatiotemporal Quantization (MASQ) framework. Our framework decouples the conflicting tasks of spatial feature inference and temporal smoothing. 
In the spatial dimension, we introduce a Joint-Level Structured Dropout (JLSD) mechanism that masks the entire temporal trajectory of selected joints, to encourage the model to learn discriminative inter-joint coordination patterns. In the temporal dimension, we design a mask-aware velocity loss that enforces motion consistency only on visible joints, that prevents gradient conflicts caused by masked signals and stabilizing temporal predictions.
Extensive experiments on three widely used skeleton datasets, including HuGaDB, LARa, and BABEL, demonstrate that the proposed MASQ framework significantly outperforms existing state-of-the-art unsupervised methods. In particular, our model establishes a comprehensive and substantial leading advantage in the Mean over Frames accuracy.
\end{abstract}

%%
%%
% Keywords for the submission system: unsupervised temporal action
% segmentation; skeleton sequences; discrete quantization; motion modeling.

%%
\section{Introduction}

\begin{figure}[t]
  \centering
  % 第一张图片（宽度设为文本宽度，上下无间距）
  % The original source referenced an unavailable seg1.jpg; the quantitative
  % comparison below is retained as the complete figure.

  % 第二张图片（紧接在第一张下方，无空白）
    \includegraphics[width=\linewidth]{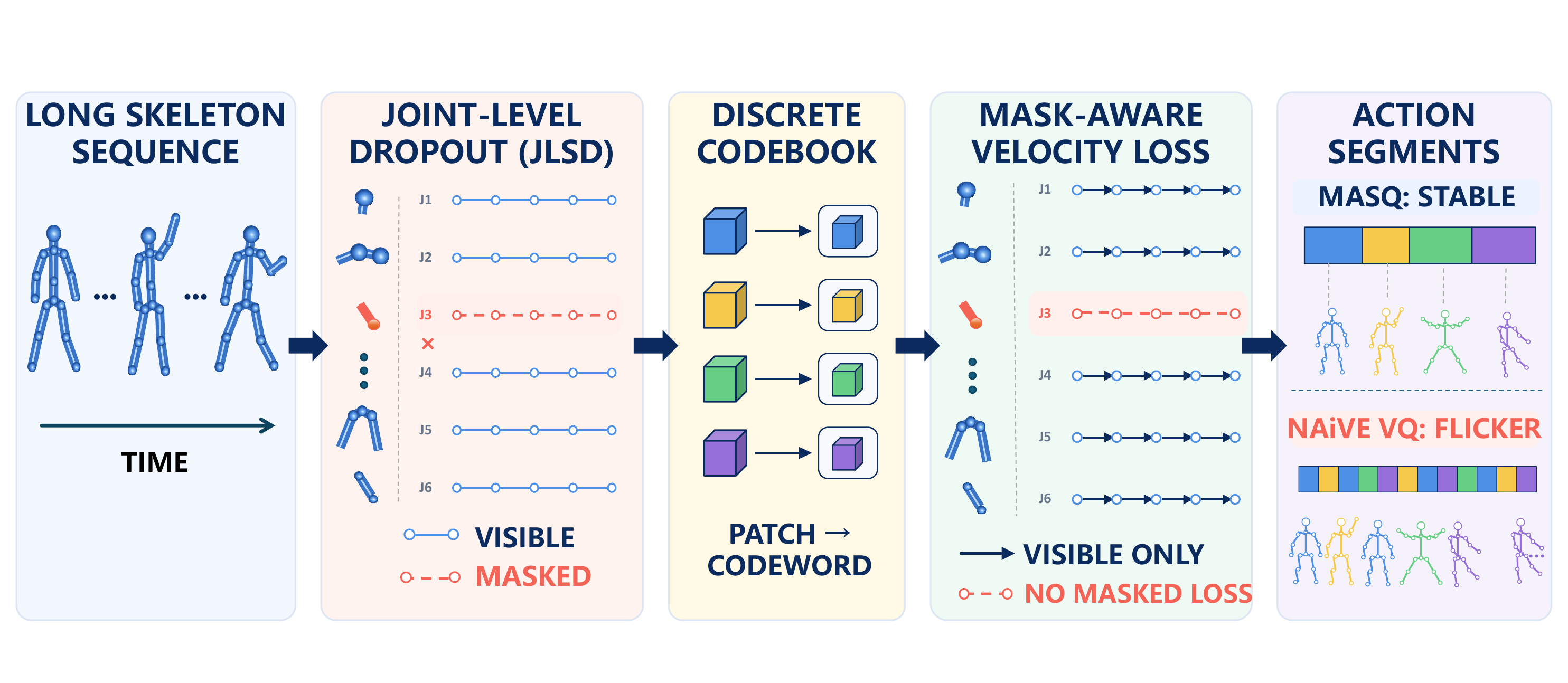}
    % 共用的标题和标签
  % \caption{\textbf{Main task of MASQ.} MASQ masks complete joint trajectories, learns discrete motion representations through vector quantization, and discovers temporally coherent action segments without supervision.Different colors correspond to distinct action categories. A small cube represents an embedding.}
  \caption{\textbf{Main task of MASQ.} JLSD masks complete joint trajectories, while the mask-aware velocity loss enforces temporal consistency only on visible joints. The learned features are quantized into discrete action tokens, producing coherent action segments and reducing the token flicker caused by naive masking and hard quantization.}
  \label{fig:compare_all}
\end{figure}

Understanding human behavior in long untrimmed sequences   \cite{farha2019ms,yi2021asformer,behrmann2022unified,li2024ms,lin2017singlest} is an important problem in multimedia analysis. Temporal action segmentation aims to divide a long sequence into a series of semantic action segments without prior knowledge of their boundaries. This capability plays a key role in many applications, including industrial process monitoring, human–robot collaboration, and healthcare analysis. 

Most existing approaches rely on fully supervised learning \cite{farha2019ms,yi2021asformer,behrmann2022unified,li2024ms}. These methods require dense frame-level annotations that specify the exact action label for every frame. Such annotations are extremely expensive and time-consuming to collect, especially for long videos. To reduce this burden, unsupervised temporal action segmentation has attracted increasing attention \cite{sener2018unsupervised,vidalmata2021joint}. Recent studies often adopt a discrete representation paradigm. They convert continuous motion signals into sequences of discrete action tokens and infer action boundaries from token transitions \cite{sedmidubsky2020motion,spurio2025hierarchical}. 

Current unsupervised segmentation methods \cite{sener2018unsupervised,vidalmata2021joint} are designed for RGB videos. However, compared with RGB videos, skeleton representations provide a structured description of human motion. They are robust to viewpoint changes, insensitive to background clutter, and naturally preserve privacy \cite{yan2018spatial,gao2019optimizedsa,gokay2025skeleton}. These properties make skeleton data an appealing modality for human behavior understanding. 
Meanwhile, skeleton-based action understanding has been extensively studied in related tasks. Recent works demonstrate that skeleton representations are highly effective for supervised action recognition \cite{yan2018spatial,gao2019optimizedsa,gokay2025skeleton}. In unsupervised learning, early studies \cite{wu2023scdnetspatiotemporalcluesdisentanglement,yang2022via} adopt contrastive learning to learn invariant motion features through data augmentation such as rotation or temporal cropping. More recently, masked reconstruction frameworks \cite{mao2023masked,yan2023skeletonmae,lin2023actionlet,thoker2021skeleton,zhang2023prompted} have shown strong performance for skeleton representation learning. These methods mask parts of the body and require the network to reconstruct the missing joints. This strategy encourages the model to capture coordination relationships among different body parts.

However, these representation-learning strategies are mainly designed for short trimmed clips. In long untrimmed sequences, masking and quantization interact with temporal structure in a way that is easy to overlook. Randomly removing joint observations at isolated times creates discontinuities that are indistinguishable from genuine transitions to an unsupervised model. Hard quantization can then amplify small latent perturbations into frequent token changes. Conversely, applying a global velocity constraint to suppress these changes imposes motion supervision on joints whose features were intentionally removed. The core problem is not masking or smoothing separately, but their conflicting handling of missing data in discrete temporal models.

% MASQ addresses this coupling by assigning the two objectives different supports. Joint-Level Structured Dropout removes complete trajectories of selected joints, so the bottleneck is spatial while the visible motion remains temporally continuous. Mask-Aware Velocity Loss applies temporal regularization only where motion evidence is available. This design encourages inter-joint inference without turning synthetic missingness into temporal boundaries or optimization targets.

To address above challenges, we propose MASQ, a novel \textbf{M}ask-aware \textbf{A}ction \textbf{S}patiotemporal \textbf{Q}uantization framework for unsupervised skeleton-based temporal action segmentation. The framework introduces two mechanisms. First, we design a Joint-Level Structured Dropout module that masks the entire temporal trajectory of randomly selected joints. This operation creates a spatial information bottleneck and encourages the model to infer the motion of masked joints from the remaining visible body parts. The design enables the network to learn stable coordination patterns across long action sequences. Second, we introduce a Mask-Aware Velocity Loss that applies temporal smoothness constraints only to visible joints. This mechanism preserves natural motion dynamics while avoiding gradient conflicts caused by masked joints. The two mechanisms work together to improve both representation robustness and temporal consistency.

% We evaluate MASQ on HuGaDB, LARa, and three BABEL subsets. Beyond comparisons with prior unsupervised methods, two analyses test the proposed interpretation. Replacing JLSD with random frame masking substantially reduces MoF, supporting the need to preserve temporal continuity. Varying the codebook size around the ground-truth class count shows that the improvements are not restricted to a single $K$. Results on BABEL further delimit the method: semantic assignment improves consistently, whereas hard quantization still limits boundary coherence under rapid transitions.

% The contributions are threefold. (i) We identify a task-specific incompatibility between random masking, global temporal regularization, and hard quantization in long untrimmed skeleton sequences. (ii) We propose a support-aware decoupling: full-trajectory JLSD creates spatial missingness, while Mask-Aware Velocity Loss regularizes only observed motion. (iii) Experiments across five evaluation sets, including masking-granularity and codebook-size analyses, validate the design choices and characterize the remaining semantic--boundary trade-off.

\begin{figure}[t]
  \centering
  % 第一张图片（宽度设为文本宽度，上下无间距）
  % The original source referenced an unavailable seg1.jpg; the quantitative
  % comparison below is retained as the complete figure.

  % 第二张图片（紧接在第一张下方，无空白）
  \includegraphics[width=\linewidth]{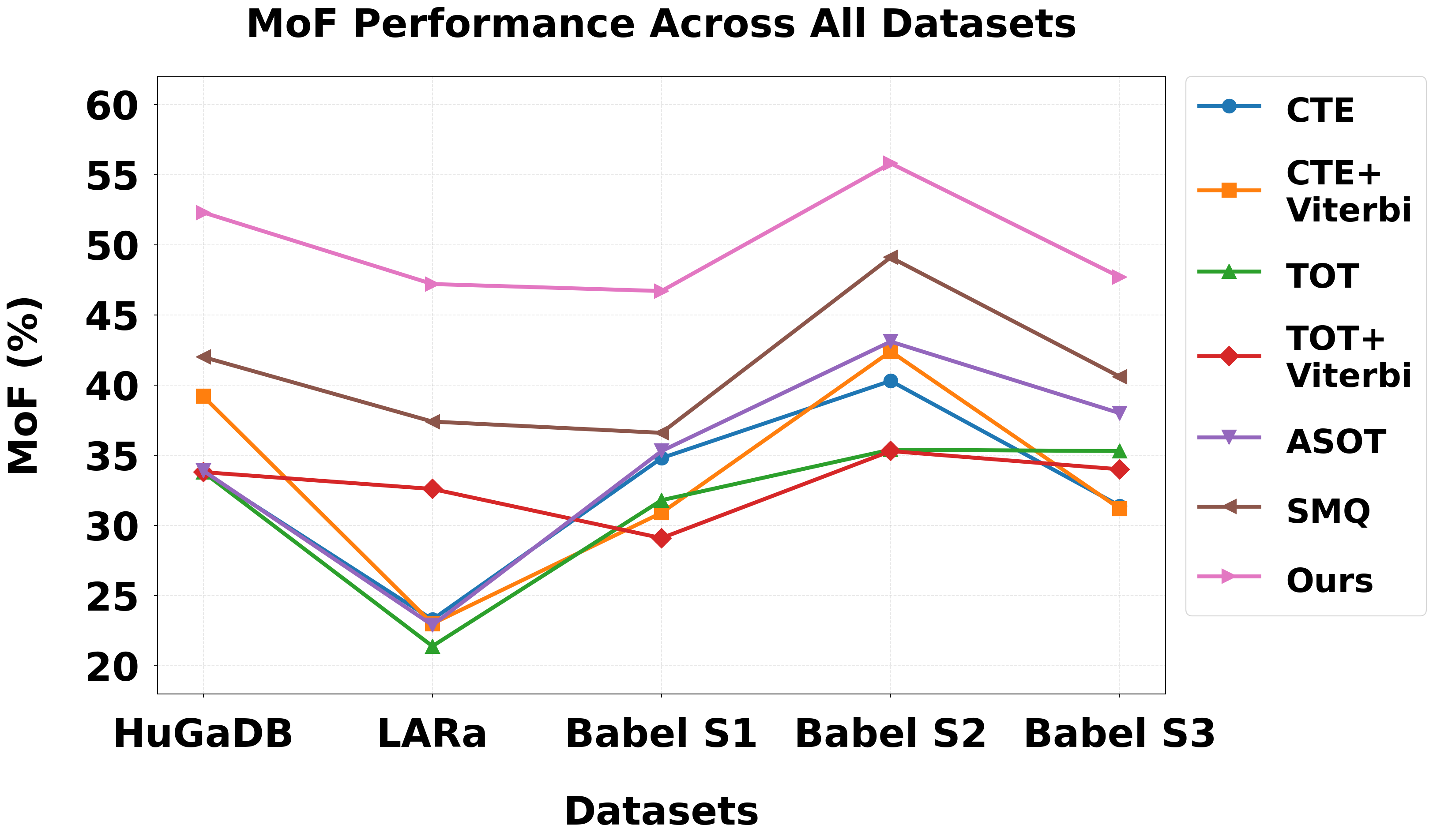}

  % 共用的标题和标签
  
  \caption{Mean over Frames (MoF) accuracy on five evaluation sets. ``S'' denotes a BABEL subset. MASQ improves MoF over the evaluated unsupervised baselines across sensor and 3D-skeleton modalities.}
  \label{fig:compare_all_lines}
\end{figure}

We evaluate the proposed framework on three widely used skeleton action datasets. Extensive experiments demonstrate that our approach consistently outperforms existing unsupervised methods across multiple evaluation metrics (see Fig.~\ref{fig:compare_all_lines}). In particular, our method achieves substantial improvements in Mean over Frames accuracy, which reflects the quality of frame-level segmentation.

The contributions of this work are summarized as follows.
(i) We propose MASQ, a novel framework for unsupervised skeleton-based temporal action segmentation that jointly addresses representation uncertainty and temporal instability in long motion sequences.
(ii) We introduce a Joint-Level Structured Dropout mechanism that masks the entire temporal trajectory of selected joints, enabling the model to capture coordination relationships among body parts and learn more robust motion representations.
(iii) We propose a Mask-Aware Velocity Loss that enforces temporal smoothness only on visible joints, which avoids conflicting supervision caused by masked joints and effectively reduces temporal jitter during segmentation.
(iv) Extensive experiments on multiple benchmark datasets demonstrate that MASQ consistently outperforms existing unsupervised methods and achieves strong frame-level segmentation performance.

\section{Related Work}

\begin{figure*}[t]  % 星号 = 跨双栏
  \centering
  \includegraphics[width=.96\textwidth]{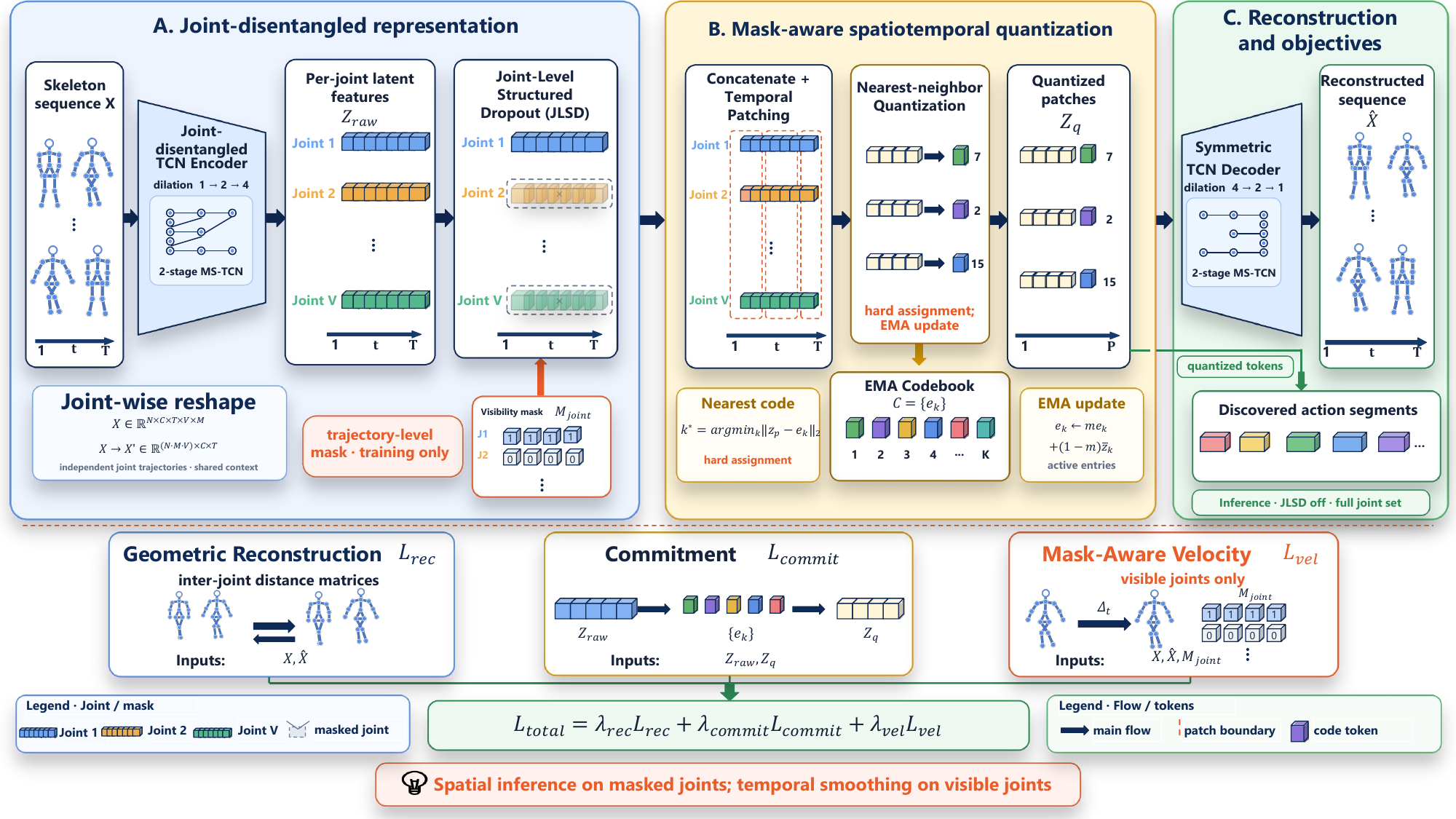}
  % \caption{MASQ overview.\textbf{ A.} A joint-disentangled encoder processes the skeleton sequence; \textbf{B.} JLSD masks complete trajectories of selected joints; temporal patches are hard-quantized using a learned codebook; \textbf{C.} and a decoder reconstructs motion. Training combines geometric reconstruction, commitment, and mask-aware velocity losses. Colors denote discovered action segments.}
\caption{Overall architecture and training objectives of MASQ.
\textbf{A. Joint-disentangled representation.} The skeleton sequence $X$ is reshaped into joint trajectories and encoded into per-joint features $Z_{\mathrm{raw}}$. During training, JLSD masks complete trajectories of selected joints.
\textbf{B. Mask-aware spatiotemporal quantization.} The features are divided into temporal patches and assigned to their nearest EMA-updated codebook entries, producing the quantized representation $Z_q$.
\textbf{C. Reconstruction and objectives.} A symmetric decoder reconstructs $\hat{X}$, while the discrete tokens support action segment discovery. Training combines geometric reconstruction loss $L_{\mathrm{rec}}$, commitment loss $L_{\mathrm{commit}}$, and mask-aware velocity loss $L_{\mathrm{vel}}$, which applies only to visible joints. JLSD is disabled during inference. Joint colors denote joint streams, and token colors denote codebook assignments and discovered action segments.}

\label{fig:overall_framework}
\end{figure*}

In this section, we review related work on temporal action segmentation and skeleton representation learning.

\subsubsection{Supervised Temporal Action Segmentation.}
Temporal action segmentation aims to divide long untrimmed sequences into distinct action segments with independent semantics. Early research in this field relies heavily on dense frame-level annotations. To capture long-range temporal dependencies, researchers have proposed various temporal convolutional networks \cite{lea2017temporal}. Recent technological advancements introduce attention mechanisms and graph convolutional networks to further model complex spatial and temporal correlations \cite{filtjens2022skeleton,li2023decoupled,hosseini2020deep}. While these supervised methods achieve excellent performance on both video and skeleton data, they require massive amounts of high-quality annotated data. This manual annotation process is extremely expensive and time-consuming, which severely restricts the application of these models in real-world large-scale scenarios. Consequently, the research community has increasingly shifted its focus toward unsupervised learning paradigms.

\subsubsection{Unsupervised Action Segmentation.}
Unsupervised action segmentation attempts to discover action boundaries without any temporal labels. Existing mainstream methods focus primarily on RGB video datasets. Some approaches utilize timestamp prediction or temporal optimal transport to generate pseudo-labels, which subsequently guide joint feature learning \cite{kukleva2019unsupervised,kumar2022unsupervised,sarfraz2021temporally}. Other studies frame the task as a representation learning and clustering problem. For instance, recent works learn features by computing temporal and semantic similarity distributions, or apply unbalanced optimal transport frameworks to obtain temporally coherent segmentation results \cite{xu2024temporally,bueno2023leveraging,sarfraz2019efficient}.

To acquire the final action segments, these video segmentation methods typically rely on Hidden Markov Models or Viterbi decoding as post-processing steps \cite{rabiner2002tutorial,richard2018neuralnetwork,tran2024permutation}. However, when these video-based methods are applied directly to skeleton sequences, their performance usually degrades significantly. They treat visual or motion features merely as generic high-dimensional vectors and fail to fully exploit the inherent geometric topology of the human skeleton \cite{du2022fast,wang2022sscap}. Furthermore, post-processing techniques relying on Markov chains often impose fixed prior assumptions on the action ordering. Such strong assumptions severely suppress the generalization ability of the models when facing flexible and diverse human actions.

\subsubsection{Skeleton Action Representation Learning.}
Unsupervised and self-supervised representation learning for skeleton data has developed rapidly in recent years. Researchers have designed various pretext tasks to learn robust action features \cite{su2020predict}. Mainstream technical branches include contrastive learning frameworks based on sequence augmentation \cite{guo2022contrastive,lin2023actionlet,zhang2022contrastive} and masked autoencoders utilizing data reconstruction \cite{zheng2018unsupervised,xu2021unsupervised}. These self-supervised methods successfully extract highly discriminative feature vectors for action recognition tasks.
Nevertheless, these methods are primarily designed for short trimmed sequences and do not distinguish synthetic missing observations from genuine transitions in long untrimmed motion. MASQ focuses on this distinction: it uses trajectory-level masking to preserve visible temporal structure and conditions temporal regularization on the same visibility mask. The model learns discrete action prototypes without frame-level labels or offline decoding.

\section{Methodology}

Figure~\ref{fig:overall_framework} shows the joint-disentangled sequence-to-sequence architecture. A temporal encoder first produces per-joint features. JLSD removes selected joint trajectories, restricting missingness to the spatial support while leaving each visible trajectory continuous. Temporal patches are then mapped to a discrete codebook and decoded to reconstruct the input motion. Training combines geometric reconstruction and codebook commitment with a velocity term evaluated only on the visible support. Thus, spatial inference and temporal regularization share the representation but not the missing entries that would make their supervision inconsistent.

\subsection{Joint-Disentangled TCN Autoencoder}

Unlike conventional frame-level masking strategies, JLSD removes the entire temporal trajectory of selected joints. This encourages the model to infer motion patterns from long-range inter-joint coordination. 
To capture fine-grained motion patterns for individual joints while maintaining global temporal context, we design an encoder and a decoder utilizing a joint-disentangled approach integrated within a Multi-Stage Temporal Convolutional Network (MS-TCN) architecture \cite{bai2018empirical}.

\textbf{Encoder :} The input skeleton sequence $X \in \mathbb{R}^{N \times C \times T \times V \times M}$ is first reshaped into $(N \cdot M \cdot V, C, T)$. 
This transformation treats each joint's temporal trajectory as an independent sample to facilitate the extraction of joint-disentangled features. 
Subsequently, these sequences are fed into a multi-stage TCN consisting of two cascaded stages. 
Each stage comprises multiple dilated residual layers where the receptive field grows exponentially with network depth, allowing for the effective capture of long-range temporal dependencies. 
The encoder finally yields high-dimensional latent embeddings $Z_{raw} \in \mathbb{R}^{(N \cdot M \cdot V) \times D_{latent} \times T}$.

\textbf{Decoder :} The decoder utilizes a network structure symmetric to that of the encoder. 
It receives the quantized features and progressively recovers the temporal resolution through a series of symmetric dilated convolutional layers. 
The final output is a reconstructed skeleton sequence $\hat{X}$ with the same dimensions as the original input.

\subsection{Joint-Level Structured Dropout}

To encourage the model to move beyond reliance on explicit local features and instead capture deep inter-joint correlations, we introduce the Joint-Level Structured Dropout (JLSD) module.

JLSD operates directly on the latent features $Z_{raw}$ produced by the encoder. 
During the training phase, a Bernoulli mask $\mathbf{M}_{joint} \in \{0, 1\}^{N \cdot M \times V}$ is generated, where each element is set to zero with a dropout probability $p$. 
This mask is broadcast across both the temporal and feature dimensions. 
Consequently, if a joint is selected for dropout, its entire temporal sequence across all feature channels is zeroed out. 
To maintain the expectation of the feature distribution during training, the remaining features are scaled by a factor of $\frac{1}{1-p}$. 
The process is formulated as:
\begin{equation}
Z_{dropped} = Z_{raw} \odot \mathbf{M}'_{joint} \times \frac{1}{1-p}
\end{equation}
where $\mathbf{M}'_{joint}$ denotes the broadcasted mask. 

JLSD requires the model to infer missing joint dynamics from observed body parts without introducing synthetic temporal boundaries. It is a training-time regularizer; inference uses the complete sequence and retains the learned inter-joint redundancy. The rate $p$ controls bottleneck strength: a small value leaves local shortcuts, whereas an overly large value removes too much motion evidence. We select $p$ using the validation protocol and analyze its sensitivity rather than assuming a universal optimum. Section~\ref{sec:ablation} directly contrasts this support choice with random frame masking.

\subsection{Temporal Patch Quantization}

After the structured dropout process, the latent features $Z_{dropped}$ are reorganized and partitioned into non-overlapping temporal windows, referred to as patches. 
These segments are subsequently passed to a discrete quantization module, where they are mapped onto a learnable codebook $\mathcal{C} = \{e_k\}_{k=1}^K$ via a nearest-neighbor assignment. 
This process effectively transforms the continuous motion representations into a sequence of prototypical action words \cite{van2017neural}, with each codebook entry representing a distinct motion pattern. 
To ensure the stability and expressiveness of the discrete embedding space, the codebook entries are updated during training using an Exponential Moving Average (EMA) scheme. 
An entry is considered inactive when it receives no assignment for a prescribed update interval. We reinitialize such an entry with a latent sample from the current training stream and resume its EMA update. This prevents codebook collapse, by which most patches concentrate on only a few codewords and distinct action semantics are merged.

\subsection{Training Objectives}

The proposed model is trained end-to-end by minimizing a joint objective function $\mathcal{L}_{total}$, defined as a weighted sum of three components:
\begin{equation}
\mathcal{L}_{total} = \lambda_{rec} \mathcal{L}_{rec} + \lambda_{commit} \mathcal{L}_{commit} + \lambda_{vel} \mathcal{L}_{vel}
\end{equation}
where $\lambda_{rec}$, $\lambda_{commit}$, and $\lambda_{vel}$ are hyperparameters that balance the contribution of each loss term.

\textbf{1. Joint Distance Reconstruction Loss ($\mathcal{L}_{rec}$):} To preserve the geometric structure and topological consistency of the reconstructed skeletons, we compute the error over inter-joint distance matrices rather than absolute coordinates. The loss is formulated as:
\begin{equation}
\mathcal{L}_{rec} = \frac{1}{N \cdot T \cdot V^2} \sum_{n,t} \| D(\hat{X}_{n,t}) - D(X_{n,t}) \|_F^2
\end{equation}
where $D(X) \in \mathbb{R}^{V \times V}$ denotes the Euclidean distance matrix between joints, with elements $D_{i,j} = \|J_i - J_j\|_2$, and $\|\cdot\|_F$ represents the Frobenius norm. This objective encourages the model to maintain consistent limb lengths and skeletal proportions.

\textbf{2. Commitment Loss ($\mathcal{L}_{commit}$):} To ensure the stability of the quantization process, we employ a commitment loss that constrains the encoder's latent features to stay close to their assigned codebook entries:
\begin{equation}
\mathcal{L}_{commit} = \mathbb{E} [\| z_e(X) - sg[e] \|_2^2]
\end{equation}
where $z_e(X)$ is the latent embedding from the encoder, $e$ is the corresponding quantized vector, and $sg[\cdot]$ denotes the stop-gradient operator.

\textbf{3. Mask-Aware Velocity Loss ($\mathcal{L}_{vel}$):} A global velocity loss would treat deliberately hidden joints as valid temporal targets, coupling spatial inference with inconsistent motion supervision. We instead compare first-order differences only on visible joints:
\begin{equation}
\begin{aligned}
\mathcal{L}_{vel} = \frac{1}{N_{vld}} \sum_{t=2}^{T} \sum_{v=1}^{V} \mathbf{1}_{m}(v)
\big\| & (\hat{X}_t^{(v)} - \hat{X}_{t-1}^{(v)}) \\
& {}- (X_t^{(v)} - X_{t-1}^{(v)}) \big\|_2^2 .
\end{aligned}
\end{equation}
where $\mathbf{1}_{m}(v)$ is an indicator function that equals 1 if the $v$-th joint is not dropped in the current batch and 0 otherwise. $N_{vld}$ is the total number of valid spatio-temporal points. This mask-aware mechanism decouples the spatial inference task from temporal smoothing, to ensure smooth motion for visible joints without introducing gradient conflicts for masked ones.
%\end{equation}
\section{Experiments}

% 表格1：HuGaDB & LARa  【已改为双栏跨栏】
\begin{table*}[t]
\centering
\small
\setlength{\tabcolsep}{2.2mm}
\begin{tabular}{c l|ccccc|ccccc}
\hline
& & \multicolumn{5}{c|}{HuGaDB} & \multicolumn{5}{c}{LARa} \\
\cline{3-12}
& Method & MoF & Edit & \multicolumn{3}{c|}{F1@\{10, 25, 50\}} & MoF & Edit & \multicolumn{3}{c}{F1@\{10, 25, 50\}} \\

\hline
& CTE  & 33.8 & 4.7 & 0.6 & 0.6 & 0.5 & 23.3 & 16.8 & 8.1 & 5.2 & 2.3 \\
& CTE + Viterbi  & 39.2 & 21.7 & 13.2 & 9.5 & 7.5 & 23.0 & 17.7 & 6.8 & 3.7 & 1.6 \\
& TOT  & 33.8 & 3.1 & 0.7 & 0.5 & 0.4 & 21.4 & 7.8 & 5.3 & 2.7 & 1.1 \\
& TOT + Viterbi  & 33.8 & 20.8 & 15.6 & 10.5 & 7.5 & 32.6 & 17.7 & 11.6 & 7.4 & 3.2 \\
& ASOT  & 33.9 & 17.4 & 4.5 & 3.8 & 3.0 & 22.9 & 23.4 & 17.8 & 12.1 & 5.7 \\
& SMQ  & 42.0 & 36.1 & 38.5 & 31.5 & 24.3 & 37.4 & 39.4 & 34.7 & 28.4 & 16.4 \\
& Ours & \textbf{52.3} & \textbf{45.1} & \textbf{52.1} & \textbf{47.0} & \textbf{39.5} & \textbf{47.2} & \textbf{42.4} & \textbf{42.0} & \textbf{35.4} & \textbf{22.2} \\
\hline
\end{tabular}
\caption{Comparison to unsupervised temporal action segmentation methods on the HuGaDB and LARa datasets.}
\label{tab:hugadb_lara}
\end{table*}

% 表格2：BABEL  【已改为双栏跨栏】
\begin{table*}[t]
\centering
\small
\setlength{\tabcolsep}{1mm}
\begin{tabular}{l|ccccc|ccccc|ccccc}
\hline
& \multicolumn{5}{c|}{BABEL Subset-1} & \multicolumn{5}{c|}{BABEL Subset-2} & \multicolumn{5}{c}{BABEL Subset-3} \\
\cline{2-16}
Method & MoF & Edit & \multicolumn{3}{c|}{F1@\{10, 25, 50\}} & MoF & Edit & \multicolumn{3}{c|}{F1@\{10, 25, 50\}} & MoF & Edit & \multicolumn{3}{c}{F1@\{10, 25, 50\}} \\
\hline
CTE  & 34.8 & 28.6 & 25.0 & 17.5 & 9.5 & 40.3 & 30.6 & 17.8 & 12.2 & 7.4 & 31.4 & 13.1 & 8.2 & 5.8 & 3.6 \\
CTE + Viterbi  & 30.9 & 36.2 & 23.2 & 15.2 & 7.3 & 42.4 & 30.7 & 24.3 & 19.5 & 12.8 & 31.2 & 30.9 & 20.7 & 15.2 & 8.4 \\
TOT  & 31.8 & 18.7 & 14.2 & 7.6 & 4.4 & 35.4 & 12.8 & 13.7 & 8.6 & 4.3 & 31.5 & 7.1 & 4.9 & 2.9 & 1.7 \\
TOT + Viterbi  & 29.1 & 29.3 & 31.5 & 20.8 & 9.9 & 35.3 & 36.8 & 35.9 & 30.0 & 19.8 & 34.0 & 33.8 & 31.3 & 26.8 & 17.9 \\
ASOT  & 35.3 & \textbf{43.1} & \textbf{42.3} & \textbf{34.1} & \textbf{24.5} & 43.1 & 37.7 & 40.3 & 33.4 & 23.4 & 38.0 & 27.1 & 27.4 & 21.6 & 14.3 \\
SMQ  & 36.6 & 38.5 & 40.9 & 32.8 & 22.3 & 49.1 & 37.8 & 43.8 & 37.4 & 27.4 & 40.6 & \textbf{38.6} & \textbf{38.0} & \textbf{29.3} & 19.3 \\
Ours  & \textbf{46.7} & 33.1 & 39.6 & 34.0 & 21.0 & \textbf{55.8} & \textbf{38.1} & \textbf{46.1} & \textbf{38.7} & \textbf{28.4} & \textbf{47.7} & 26.9 & 28.3 & 27.5 & \textbf{24.5} \\

\hline
\end{tabular}
\caption{Comparison to unsupervised temporal action segmentation methods on the BABEL dataset.}
\label{tab:babel}
\end{table*}

In this section, we present the quantitative and qualitative evaluation of the MASQ framework.

\subsection{Datasets}
We validate our framework on three standard skeleton-based action datasets: HuGaDB, LARa, and BABEL. HuGaDB \cite{chereshnev2017hugadb} includes 364 trials of 10 lower-limb activities captured via IMUs. LARa \cite{niemann2020lara} provides 439 trials of 8 warehouse actions with full-body 22-joint motion capture data, which we downsample to 50 fps and root-center for translation invariance. BABEL \cite{punnakkal2021babel,mahmood2019amass} is a large-scale dataset with 43 hours of AMASS-derived 3D motion, over 63k frame labels across 250+ action classes; following standard protocols, we extract 25 full-body joints, build three four-class subsets, downsample sequences to 30 fps, root-center skeletons, and discard clips with over 50\% irrelevant background motions.

\subsection{Evaluation Metrics}

To comprehensively measure the performance of our proposed framework on the temporal action segmentation task, we employ three standard evaluation metrics widely used in this field. These metrics include the Mean over Frames (MoF) accuracy, the segmental Edit Score, and the segmental F1 scores \cite{lea2017temporal} at overlapping thresholds of 10\%, 25\%, and 50\%.

The Edit Score evaluates the accuracy of the action ordering by calculating the Levenshtein distance between the predicted and ground truth segment sequences. The F1 scores further measure the precision and recall of the predicted action boundaries. These two segmental metrics jointly reflect the global coherence of the segmentation results and strictly penalize over-segmentation phenomena.

MoF measures frame-level semantic assignment, whereas Edit and F1 additionally expose ordering, fragmentation, and boundary quality. We report all metrics without treating one as decisive: MoF is particularly informative for unsupervised semantic discovery, while Edit/F1 reveal whether that accuracy is obtained with coherent segments.

% \textbf{Implementation details.} The encoder and decoder use two MS-TCN stages with three dilated residual layers per stage, 128 feature maps, and a 16-dimensional latent vector per joint. One-second patches contain 60, 50, and 30 frames for HuGaDB, LARa, and BABEL, respectively. We train for 30 epochs with Adam (learning rate $5\times10^{-4}$), EMA codebook decay 0.5, reconstruction/commitment/velocity weights $0.001/1.0/0.001$, and batch sizes 8 (HuGaDB/LARa) or 32 (BABEL). JLSD rates are 0.25, 0.20, and 0.50 for HuGaDB, LARa, and BABEL. All runs use one NVIDIA RTX A6000 GPU. Additional sensitivity curves and qualitative examples are provided in the supplementary material.

\subsection{Comparison with State of the Art}

\begin{figure*}[t]
    \centering
    % 第一行
    \begin{subfigure}{0.48\textwidth}
        \centering
        \includegraphics[width=\linewidth]{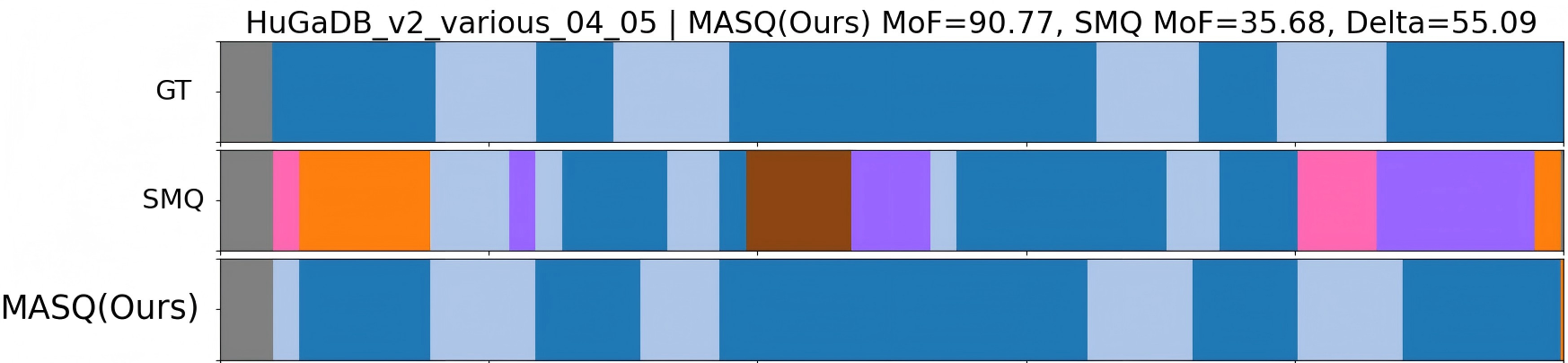}
        \caption{HuGaDB}
    \end{subfigure}
    \hfill
    \begin{subfigure}{0.48\textwidth}
        \centering
        \includegraphics[width=\linewidth]{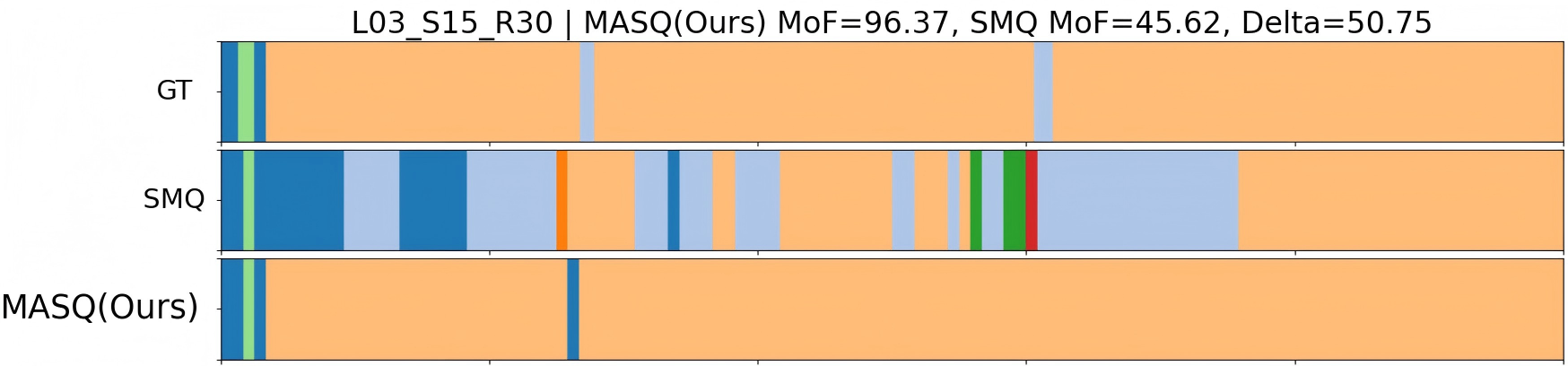}
        \caption{LARa}
    \end{subfigure}
    \\[10pt] % 换行 + 间距
    % 第二行
    \begin{subfigure}{0.48\textwidth}
        \centering
        \includegraphics[width=\linewidth]{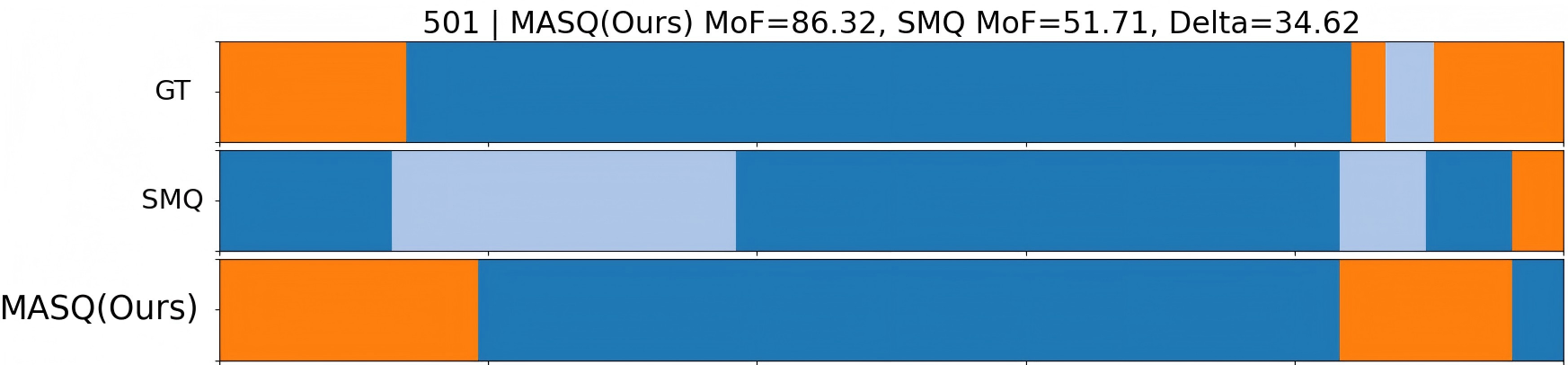}
        \caption{BABEL Subset-2}
    \end{subfigure}
    \hfill
    \begin{subfigure}{0.48\textwidth}
        \centering
        \includegraphics[width=\linewidth]{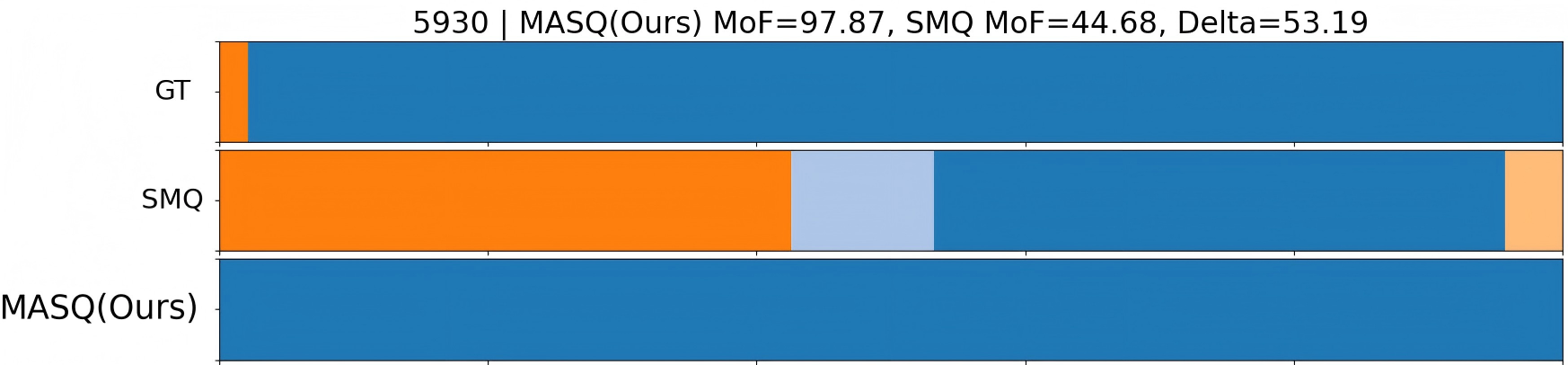}
        \caption{BABEL Subset-3}
    \end{subfigure}
    \caption{Representative frame-level segmentations. MASQ better preserves dominant action regions than SMQ, while the BABEL examples retain some boundary offsets and short fragments. Sequence-level MoF is shown in each panel.}
    \label{fig:qualitative_results}
\end{figure*}

% We compare MASQ with unsupervised temporal action segmentation methods, including CTE \cite{kukleva2019unsupervised}, TOT \cite{kumar2022unsupervised}, ASOT \cite{xu2024temporally}, and the discrete quantization baseline SMQ \cite{gokay2025skeleton}. Fully supervised methods use dense frame labels and are therefore annotation-heavy upper bounds rather than fair baselines. Tables~\ref{tab:hugadb_lara} and~\ref{tab:babel} report the unsupervised comparison.

We compare MASQ against unsupervised temporal action segmentation approaches: CTE \cite{kukleva2019unsupervised}, TOT \cite{kumar2022unsupervised}, ASOT \cite{xu2024temporally}, and discrete quantization baseline SMQ \cite{gokay2025skeleton}. Fully supervised models rely on dense frame annotations, serving as annotation-costly upper bounds instead of equitable baselines. Tables~\ref{tab:hugadb_lara} and~\ref{tab:babel} present this unsupervised evaluation.

As shown in Table~\ref{tab:hugadb_lara}, our method achieves substantial gains in MoF, Edit Score, and all F1 scores on both HuGaDB and LARa. On the HuGaDB dataset, our model reaches a MoF of 52.3\%, surpassing the state-of-the-art baseline by 10.3 percentage points. Similarly, on the LARa dataset, the MoF increases from 37.4\% to 47.2\%. The significant improvement in F1@50 scores indicates an effective synergy between the joint-level structured dropout and the mask-aware velocity loss. This combination enables the model to generate more precise and kinematically consistent action segments.

Table~\ref{tab:babel} evaluates three BABEL subsets. MASQ obtains the highest MoF on each subset (46.7\%, 55.8\%, and 47.7\%), but its Edit/F1 gains are not uniform. It leads all reported metrics on Subset-2 and F1@50 on Subset-3, while ASOT or SMQ remains stronger on several boundary-sensitive measures. MASQ boosts semantic action discovery yet struggles with boundary localization under fast complex 3D motions.

On HuGaDB and LARa, the segment-sensitive improvements indicate that MASQ reduces fragmentation as well as frame errors. On BABEL, the mixed Edit/F1 results show that code switching can persist around rapid transitions; we therefore avoid inferring boundary quality from MoF alone.

Figure~\ref{fig:qualitative_results} complements the aggregate scores. MASQ removes many short label oscillations on HuGaDB and LARa and recovers the dominant regions in both BABEL examples. Residual BABEL fragments and shifted endpoints nevertheless remain visible, consistent with the mixed Edit/F1 results.

\begin{table*}[t]
\centering

\small
\setlength{\tabcolsep}{2.2mm}
\begin{tabular}{l | c c c c c | c c c c c}
\hline
& \multicolumn{5}{c|}{HuGaDB} & \multicolumn{5}{c}{LARa} \\
Variant & MoF & Edit & F1@10 & 25 & 50 & MoF & Edit & F1@10 & 25 & 50 \\
\hline
+ JL Only & 50.8 & 34.2 & 42.0 & 37.4 & 31.0 & 41.4 & \textbf{42.5} & 41.2 & 34.3 & 20.3 \\
+ Vel. Only & 44.2 & 39.4 & 41.9 & 33.7 & 25.7 & 37.3 & 39.0 & 34.8 & 28.4 & 16.1 \\
+ Global Vel. & 50.0 & 43.1 & 45.6 & 39.2 & 35.1 & 44.2 & 41.5 & 38.7 & 32.2 & 19.2 \\
\hline
\textbf{Ours(full)} & \textbf{52.3} & \textbf{45.1} & \textbf{52.1} & \textbf{47.0} & \textbf{39.5} & \textbf{47.2} & 42.4 & \textbf{42.0} & \textbf{35.4} & \textbf{22.2} \\
\hline
\end{tabular}
\caption{Component ablation on HuGaDB and LARa. ``JL'' denotes JLSD; ``Global Vel.'' also regularizes masked joints.}
\label{tab:ablation_comprehensive_1}
\end{table*}

\begin{table*}[t]
\centering

\small
\setlength{\tabcolsep}{1mm}
\begin{tabular}{l | c c c c c | c c c c c | c c c c c}
\hline
& \multicolumn{5}{c|}{BABEL Subset 1} & \multicolumn{5}{c|}{BABEL Subset 2} & \multicolumn{5}{c}{BABEL Subset 3} \\
Variant & MoF & Edit & F1@10 & 25 & 50 & MoF & Edit & F1@10 & 25 & 50 & MoF & Edit & F1@10 & 25 & 50 \\
\hline
+ JL Only & 44.7 & \textbf{35.9} & \textbf{42.3} & \textbf{36.7} & \textbf{22.9} & \textbf{56.3} & \textbf{38.8} & \textbf{48.2} & \textbf{42.5} & \textbf{34.0} & 43.3 & 28.9 & \textbf{31.2} & 23.2 & 17.8 \\
+ Vel. Only & 36.4 & 26.3 & 31.0 & 25.1 & 12.9 & 47.3 & 36.0 & 33.4 & 27.4 & 18.2 & 41.3 & \textbf{37.4} & 36.4 & \textbf{29.0} & 19.5 \\
+ Global Vel. & 38.6 & 28.9 & 36.0 & 28.5 & 13.2 & 44.6 & 30.3 & 32.8 & 27.4 & 17.8 & 41.1 & 27.3 & 23.1 & 19.2 & 14.5 \\
\hline
\textbf{Ours (Full)} & \textbf{46.7} & 33.1 & 39.6 & 34.0 & 21.0 & 55.8 & 38.1 & 46.1 & 38.7 & 28.4 & \textbf{47.7} & 26.9 & 28.3 & 27.5 & \textbf{24.5} \\
\hline
\end{tabular}
\caption{Component ablation on the three BABEL subsets. ``JL'' denotes JLSD; ``Global Vel.'' also regularizes masked joints.}
\label{tab:ablation_comprehensive_2}
\end{table*}

\begin{table}[t]
\centering
\small
\setlength{\tabcolsep}{2.mm}
\begin{tabular}{llccccc}
\toprule
Dataset & Mask & MoF & Edit & F1@10 & 25 & 50 \\
\midrule
\multirow{2}{*}{HuGaDB}
 & Frame & 45.5 & 43.1 & 46.3 & 36.8 & 26.4 \\
 & JLSD  & \textbf{52.3} & \textbf{45.1} & \textbf{52.1} & \textbf{47.0} & \textbf{39.5} \\
\midrule
\multirow{2}{*}{LARa}
 & Frame & 32.8 & 38.6 & 33.8 & 26.6 & 14.9 \\
 & JLSD  & \textbf{47.2} & \textbf{42.4} & \textbf{42.0} & \textbf{35.4} & \textbf{22.2} \\
\midrule
\multirow{2}{*}{BABEL-1}
 & Frame & 36.3 & 30.3 & 34.4 & 27.5 & 14.5 \\
 & JLSD  & \textbf{46.7} & \textbf{33.1} & \textbf{39.6} & \textbf{34.0} & \textbf{21.0} \\
\midrule
\multirow{2}{*}{BABEL-2}
 & Frame & 50.0 & \textbf{39.0} & 43.2 & 36.5 & 25.7 \\
 & JLSD  & \textbf{55.8} & 38.1 & \textbf{46.1} & \textbf{38.7} & \textbf{28.4} \\
\midrule
\multirow{2}{*}{BABEL-3}
 & Frame & 36.4 & 25.5 & 24.6 & 17.7 & 11.0 \\
 & JLSD  & \textbf{47.7} & \textbf{26.9} & \textbf{28.3} & \textbf{27.5} & \textbf{24.5} \\
\bottomrule
\end{tabular}
\caption{Masking-granularity comparison. Frame masking uses its fixed reported configuration ($p{=}0.25$, $n{=}10$); JLSD uses the dataset-specific configurations in the main comparison.}
\label{tab:masking_granularity}
\end{table}

\begin{table}[t]
\centering
\small
\setlength{\tabcolsep}{2.5mm}
\begin{tabular}{llccccc}
\toprule
Dataset & $K$ & MoF & Edit & F1@10 & 25 & 50 \\
\midrule
\multirow{5}{*}{HuGaDB}
 & 5  & 44.4 & 34.3 & 35.8 & 28.4 & 19.4 \\
 & 8  & 50.8 & 42.0 & 49.7 & 45.6 & 37.6 \\
 & 10 & \textbf{52.3} & \textbf{45.1} & \textbf{52.1} & \textbf{47.0} & \textbf{39.5} \\
 & 12 & 48.7 & 44.8 & 49.1 & 38.8 & 31.1 \\
 & 15 & 51.3 & 38.4 & 44.8 & 37.7 & 29.3 \\
\midrule
\multirow{5}{*}{LARa}
 & 4  & 42.7 & 32.5 & 31.9 & 24.7 & 13.9 \\
 & 6  & 43.4 & 37.8 & 37.7 & 30.2 & 17.0 \\
 & 8  & \textbf{47.2} & \textbf{42.4} & \textbf{42.0} & \textbf{35.4} & \textbf{22.2} \\
 & 10 & 38.6 & 40.7 & 36.8 & 29.7 & 17.6 \\
 & 12 & 41.3 & 39.0 & 38.3 & 31.2 & 17.3 \\
\bottomrule
\end{tabular}
\caption{Codebook-size sensitivity. $K_0{=}10$ for HuGaDB and $K_0{=}8$ for LARa; actual integer sizes are reported.}
\label{tab:codebook_sensitivity}
\end{table}

\subsection{Ablation Studies}\label{sec:ablation}
%To comprehensively validate the effectiveness of the individual components across various evaluation dimensions, 
% We first isolate the two design choices behind the support-aware formulation. Tables~\ref{tab:ablation_comprehensive_1} and~\ref{tab:ablation_comprehensive_2} compare JLSD, velocity regularization, their global coupling, and the complete mask-aware model. These tables report a controlled component-ablation run, whereas Table~\ref{tab:masking_granularity} uses the main-comparison configuration; this distinction accounts for small run-level differences such as the BABEL-2 full-model values.
We perform ablation studies to verify the efficacy of JLSD and mask-aware velocity loss, and further investigate masking strategies and codebook size sensitivity.

\textbf{Component interaction.} JLSD alone provides most of the semantic gain, but the mask-aware velocity term is important on the sensor-based datasets: adding the complete formulation raises MoF from 50.8 to 52.3 on HuGaDB and from 41.4 to 47.2 on LARa, with corresponding F1@50 gains. Applying velocity supervision globally is consistently weaker than excluding masked joints, supporting the proposed separation of spatial inference from temporal supervision. BABEL is less uniform: JLSD alone can retain stronger segmental scores on Subset-2, whereas the full model improves the reported main-run MoF across all three subsets. This reinforces the narrower conclusion that mask awareness prevents invalid gradients, while a first-order smoothness prior cannot guarantee optimal boundaries under hard quantization.

\textbf{Masking granularity.} Table~\ref{tab:masking_granularity} directly contrasts random temporal masking with full-trajectory JLSD. JLSD improves MoF by 6.8 and 14.4 points on HuGaDB and LARa and by 5.8--11.3 points on BABEL. The gains extend to F1@50 on all five sets, supporting the claim that preserving the temporal continuity of visible joints is important. BABEL-2 Edit decreases slightly (39.0 to 38.1), so the experiment supports semantic discovery and overlap quality without implying uniform boundary improvement. This comparison assesses full masking configurations instead of tuning $p$ alone

\textbf{Codebook sensitivity.} Table~\ref{tab:codebook_sensitivity} varies the codebook resolution around the class-matched $K_0$. MASQ remains above the SMQ MoF at $K_0$ (42.0 on HuGaDB and 37.4 on LARa) for every tested $K$, including its lowest values of 44.4 and 38.6. The best results still occur at $K_0$, and the decline for larger LARa codebooks cautions against claiming codebook-size invariance. The sweep instead shows that the improvement is not confined to one exact quantization resolution.

\textbf{Strength of the spatial bottleneck.} The supplementary analysis varies the JLSD rate from 0.20 to 0.50. HuGaDB and LARa peak at 0.25 and 0.20, respectively, whereas the three BABEL subsets favor 0.50. This pattern is consistent with the role assigned to JLSD: repetitive or lower-dimensional motions require only a mild bottleneck to remove local shortcuts, while higher-dimensional BABEL motions benefit from stronger inter-joint inference. Performance degrades when too much evidence is removed from HuGaDB or LARa, so the results do not support a universally aggressive masking rate. Instead, $p$ controls how much spatial evidence is retained without changing the temporal support of the mask.

\textbf{Strength of temporal regularization.} Supplementary sweeps vary the mask-aware velocity weight over several orders of magnitude. A weight of $0.001$ provides the best overall balance: very small weights insufficiently constrain local fluctuations, whereas large weights over-smooth transitions. Together with the global-velocity ablation, this separates invalid supervision on masked support from excessive smoothing on valid support.

\textbf{Computational overhead.} JLSD is an element-wise training mask and the velocity term uses first-order temporal differences only during training. Neither changes the encoder, decoder, or nearest-neighbor assignment at inference; consequently, MASQ has the same inference architecture and asymptotic cost as the SMQ backbone, with only linear-time training regularization overhead.

\textbf{Continuous smoothing versus hard assignment.} Velocity regularization acts in a continuous feature space, but nearest-neighbor quantization is discontinuous. Features close to a Voronoi boundary can therefore alternate between adjacent codes even when their trajectories are smooth. Such switches affect relatively few frames and have limited influence on MoF, yet they create extra segments that are heavily penalized by Edit and F1. On the more periodic HuGaDB and LARa motions, temporal regularization mainly suppresses local noise and improves both frame- and segment-level metrics. On BABEL, rapid semantic transitions make boundary-adjacent assignments more common. We thus attribute the results to a semantic-boundary trade-off, not proof that MASQ fully solves boundary localization.

\textbf{Suboptimal Uniformity of Full on BABEL} BABEL has faster, more heterogeneous transitions than HuGaDB and LARa. Full still improves JLSD-only MoF by 2.0 and 4.4 points on Subsets 1 and 3, respectively, and raises Subset-3 F1@50 from 17.8 to 24.5, but JLSD-only is stronger on most boundary metrics and on Subset-2 MoF. Mask awareness removes erroneous gradients from hidden joints, yet cannot distinguish a visible within-action frame from a visible true boundary. The first-order prior may therefore suppress legitimate velocity discontinuities; hard quantization can convert the resulting boundary-adjacent trajectory into code switches. This trade-off explains higher semantic accuracy without uniform Edit/F1 gains and identifies boundary-agnostic smoothing, rather than JLSD, as the limitation.

% \textbf{Experimental scope.} The masking comparison does not isolate masking probability, and the codebook sweep covers HuGaDB and LARa but not BABEL. We therefore treat them as evidence for masking structure and robustness to several quantization resolutions, not as probability-controlled or domain-universal claims.

\section{Conclusion}

We presented MASQ, a support-aware framework for unsupervised skeleton action segmentation. Full-trajectory JLSD removes entire joint trajectories without introducing artificial temporal gaps, while the mask-aware velocity loss regularizes only visible joints. Across HuGaDB, LARa, and BABEL, MASQ improves frame-level action discovery and often reduces fragmentation. Ablations confirm that the gains are not tied to frame masking or a single codebook size. Results on BABEL also reveal the remaining limitation: hard nearest-neighbor quantization can still cause boundary jitter near rapid transitions. Future work will explore softer assignments or explicit boundary models to improve Edit and F1 while preserving frame-level discrimination.
%\clearpage
%\newpage

% Add acknowledgments here if appropriate for the public arXiv version.

%\clearpage
\bibliography{ref}

% \appendix









\end{document}